\documentclass[sigconf]{acmart}

\setcopyright{none}
\renewcommand\footnotetextcopyrightpermission[1]{}
\AtBeginDocument{%
  \providecommand\BibTeX{{%
    \normalfont B\kern-0.5em{\scshape i\kern-0.25em b}\kern-0.8em\TeX}}}

\usepackage{algorithm}
\usepackage[noend]{algpseudocode}
\usepackage{xcolor}
\usepackage{multirow}
\usepackage{makecell}

\usepackage{microtype}

\DeclareMathOperator*{\argmax}{arg\,max}

\newcommand{\algonameshort}{DCG-MAP-Elites}
\newcommand{\algonamelong}{Descriptor-Conditioned Gradients MAP-Elites}
\newcommand{\cparam}{\theta}
\newcommand{\aparam}{\phi}
\newcommand{\sparam}{\psi}
\newcommand{\ncritic}{n}
\newcommand{\npg}{m}

\usepackage{enumitem}

\newcommand{\delay}{\Delta}

\begin{document}

\title{MAP-Elites with Descriptor-Conditioned Gradients and Archive~Distillation into a Single Policy}

\author{Maxence Faldor}
\email{m.faldor22@imperial.ac.uk}
\affiliation{%
  \institution{Imperial College London}
  \city{London}
  \country{United Kingdom}
}

\author{Félix Chalumeau}
\email{f.chalumeau@instadeep.com}
\affiliation{%
  \institution{InstaDeep}
  \city{Cape Town}
  \country{South Africa}
}

\author{Manon Flageat}
\email{manon.flageat18@imperial.ac.uk}
\affiliation{%
  \institution{Imperial College London}
  \city{London}
  \country{United Kingdom}
}

\author{Antoine Cully}
\email{a.cully@imperial.ac.uk}
\affiliation{%
  \institution{Imperial College London}
  \city{London}
  \country{United Kingdom}
}

\renewcommand{\shortauthors}{Faldor et al.}

\begin{abstract}

Quality-Diversity algorithms, such as MAP-Elites, are a branch of Evolutionary Computation generating collections of diverse and high-performing solutions, that have been successfully applied to a variety of domains and particularly in evolutionary robotics. However, MAP-Elites performs a divergent search based on random mutations originating from Genetic Algorithms, and thus, is limited to evolving populations of low-dimensional solutions. PGA-MAP-Elites overcomes this limitation by integrating a gradient-based variation operator inspired by Deep Reinforcement Learning which enables the evolution of large neural networks. Although high-performing in many environments, PGA-MAP-Elites fails on several tasks where the convergent search of the gradient-based operator does not direct mutations towards archive-improving solutions. In this work, we present two contributions: (1) we enhance the Policy Gradient variation operator with a descriptor-conditioned critic that improves the archive across the entire descriptor space, (2) we exploit the actor-critic training to learn a descriptor-conditioned policy at no additional cost, distilling the knowledge of the archive into one single versatile policy that can execute the entire range of behaviors contained in the archive. Our algorithm, \algonameshort{} improves the QD score over PGA-MAP-Elites by 82\% on average, on a set of challenging locomotion tasks.

\end{abstract}





\maketitle

\pagestyle{plain}

\begin{figure}[ht]
    \label{fig:banner-short}
    \centering
    \includegraphics[width=\hsize]{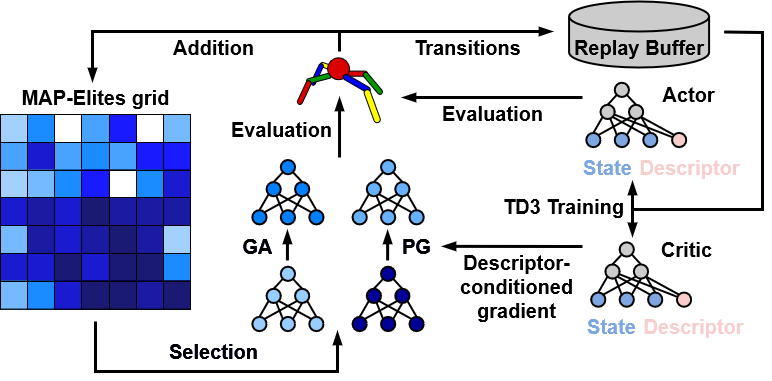}
    \caption{\algonameshort{} performs a standard MAP-Elites loop of selection, variation, evaluation and addition. Two complementary variation operators are applied: (1) a standard Genetic Algorithm (GA) variation operator for exploration, (2) a Descriptor-Conditioned Policy Gradient (PG) variation operator for fitness improvement. Concurrently to the critic's training, the knowledge of the archive is distilled in the descriptor-conditioned actor as by-product.}
\end{figure}

\section{Introduction}

A fascinating aspect of evolution is its ability to generate a large variety of different species, each being adapted to their ecological niche. Inspired by this idea, Quality-Diversity (QD) optimization is a family of evolutionary algorithms that aims to generate a set of both diverse and high-performing solutions to a problem~\cite{chatzilygeroudis_quality-diversity_2020,cully_quality_2017,pugh_quality_2016}. Contrary to traditional optimization methods that return a single high-performing solution, the goal of QD algorithms is to illuminate a search space of interest called \emph{descriptor space}~\cite{mouret_illuminating_2015}. Producing a large collection of diverse and effective solutions enables to get multiple alternatives to solve a single problem which is useful in robotics to improve robustness, recover from damage~\cite{cully_robots_2015} or reduce the reality gap \cite{chatzilygeroudis_reset-free_2018}. Furthermore, conventional optimization methods are prone to get stuck in local optima whereas keeping diverse solutions to a problem can help to find stepping stones that lead to globally better solutions~\cite{mouret_illuminating_2015,nilsson_policy_2021}. Another benefit of diversity search is efficient exploration in problems where the reward signal is sparse or deceptive~\cite{ecoffet_go-explore_2021,chalumeau_assessing_2022,pierrot_diversity_2022}.

MAP-Elites~\cite{mouret_illuminating_2015} is a conceptually simple but effective QD optimization algorithm that has shown competitive results in a variety of applications, to generate large collections of diverse skills. However, MAP-Elites relies on random variations that can cause slow convergence in large search space~\cite{colas_scaling_2020,nilsson_policy_2021,pierrot_diversity_2022}, making it inadequate to evolve neural networks with a large number of parameters.

In contrast, Deep Reinforcement Learning (DRL)~\cite{mnih_playing_2013,mnih_human-level_2015} algorithms combine reinforcement learning with the directed search power of gradient-based methods in order to learn a single solution. DRL can surpass human performance at video games~\cite{vinyals_grandmaster_2019}, beat world champions in board games~\cite{silver_mastering_2016} and control complex robots in continuous action spaces~\cite{gu_deep_2016}, which is a long-standing challenge in artificial intelligence. Policy Gradient methods have shown state-of-the-art results to learn large neural network policies with thousands of parameters in high-dimensional state space and continuous action space~\cite{lillicrap_continuous_2019,silver_deterministic_2014,haarnoja_soft_2018}.

PGA-MAP-Elites~\cite{nilsson_policy_2021} is an extension of MAP-Elites that integrates the sample efficiency of DRL using the TD3 algorithm~\cite{fujimoto_addressing_2018}. This algorithm uses a Policy Gradient (PG) variation operator for efficient fitness improvement, coupled with the usual Genetic Algorithm (GA) variation operator. The PG variation operator leverages gradients derived from DRL to improve fitness and drive mutations towards the global optimum and is supported by the divergent search of the GA variation operator for both exploration and optimization~\cite{flageat_empirical_2022}. Other recent works have also introduced methods to combine the strength of QD algorithms with reinforcement learning~\cite{tjanaka_approximating_2022,pierrot_diversity_2022} on complex robotics tasks.

PGA-MAP-Elites achieves state-of-the-art performances in most of the environments considered so far in the literature~\cite{nilsson_policy_2021,pierrot_diversity_2022,tjanaka_approximating_2022}. However, the PG variation operator becomes ineffective in tasks where the global optimum is in an area of the search space that is not likely to produce offspring that are added to the archive. For example, consider a locomotion task where the fitness is the opposite of the energy consumption and the descriptor is defined as the final position of the robot. The global optimum for the fitness is the solution that does not move in order to minimize energy consumption. Thus, the PG variation operator will encourage solutions to stay motionless, collapsing their descriptors to a single point, the descriptor of the global optimum. Consequently, the PG variation operator generates offspring that are discarded and no interesting stepping stone is found, thereby hindering diversity.

In this work, we introduce \algonamelong{} (\algonameshort{}) that builds upon PGA-MAP-Elites algorithm by enhancing the PG variation operator with a descriptor-conditioned critic that provides gradients depending on a target descriptor. The descriptor-conditioned critic takes as input a state and a descriptor to evaluate actions. With such a descriptor-conditioned critic, the PG variation operator can mutate solutions to produce offspring with higher fitness while targeting a desired descriptor, thereby avoiding to collapse the descriptor to a single point.

Furthermore, TD3 which is the DRL algorithm used by the PG variation operator, requires to train an actor and a critic in parallel. We take advantage of this intertwined actor-critic training to make the actor descriptor-conditioned as well, allowing it to take actions based on the current state and on an input descriptor we want to achieve. Thus, instead of taking actions that maximize the fitness globally, the actor now takes actions that maximize the fitness while achieving a desired descriptor. At the end of training, we can condition the actor on a desired descriptor to execute a policy that takes actions that achieve the desired descriptor. On half of the tasks, we observe that the descriptor-conditioned actor can achieve the entire range of descriptors contained in the archive with a similar QD-score, negating the burden of dealing with a collection of thousands of solutions.

In summary, we present two contributions: (1) we enhance the PG variation operator with a descriptor-conditioned critic, (2) we distill the knowledge of the archive into one single versatile policy at no additional cost. We compare our algorithm to four state-of-the-art QD algorithms on four high-dimensional robotics locomotion tasks. The results demonstrate that \algonameshort{} has a QD-score 82\% higher than PGA-MAP-Elites on average.

\section{Background}

\subsection{Problem Statement}

We consider an agent sequentially interacting with an environment at discrete time steps $t$ for an episode of length $T$. At each time step $t$, the agent observes a state $s_t$, takes an action $a_t$ and receives a scalar reward $r_t$. We model it as a Markov Decision Process (MDP) which comprises a \emph{state space} $\mathcal{S}$, a continuous \emph{action space} $\mathcal{A}$, a stationary \emph{transition dynamics distribution} $p(s_{t+1} \mid s_t, a_t)$ and a \emph{reward function} $r \colon \mathcal{S} \times \mathcal{A} \to \mathbb{R}$. In this work, a \emph{policy} (also called \emph{solution}) is a deterministic neural network parameterized by $\aparam \in \Phi$, and denoted $\pi_\aparam \colon \mathcal{S} \to \mathcal{A}$. The agent uses its policy to select actions and interact with the environment to give a trajectory of states, actions and rewards. The \emph{fitness} of a solution is given by $F \colon \Phi \to \mathbb{R}$, defined as the expected discounted return $\mathbb{E}_{\pi_\aparam} \left[ \sum_{t=0}^{T-1} \gamma^t r_t \right]$.

The objective of QD algorithms in this MDP setting is to find the highest-fitness solutions in each point of the \emph{descriptor space} $\mathcal{D}$. The descriptor function $D \colon \Phi \to \mathcal{D}$ is generally defined by the user and characterize solutions in a meaningful way for the type of diversity desired. With this notation, our objective is to evolve a population of solutions that are both high-performing with respect to $F$ and diverse with respect to $D$.

\subsection{MAP-Elites}

Multi-dimensional Archive of Phenotypic Elites (MAP-Elites)~\cite{mouret_illuminating_2015} is a simple yet effective QD algorithm that discretizes the descriptor space $\mathcal{D}$ into a multi-dimensional grid of cells called archive $\mathcal{X}$ and searches for the best solution in each cell, see Alg.~\ref{alg:map-elites}. The goal of the algorithm is to return an archive that is filled as much as possible with high-fitness solutions. MAP-Elites starts by generating random solutions and adding them to the archive. The algorithm then repeats the following steps until a budget of $I$ solutions have been evaluated: (1) a batch of solutions from the archive are uniformly selected and modified through mutations and/or crossovers to produce offspring, (2) the fitnesses and descriptors of the offspring are evaluated, and each offspring is placed in its corresponding cell if and only if the cell is empty or if the offspring has a better fitness than the current solution in that cell, in which case the current solution is replaced. As most evolutionary methods, MAP-Elites relies on undirected updates that are agnostic to the fitness objective. With a Genetic Algorithm (GA) variation operator, MAP-Elites performs a divergent search that may cause slow convergence in high-dimensional problems due to a lack of directed search power, and thus, is performing best on low-dimensional search space \cite{nilsson_policy_2021}.

\subsection{Deep Reinforcement Learning}
\label{sec:drl}

Deep Reinforcement Learning (DRL)~\cite{mnih_playing_2013,mnih_human-level_2015} combines the reinforcement learning framework with the function approximation capabilities of deep neural networks to represent policies and value functions in high-dimensional state and action spaces. In opposition to black-box optimization methods like evolutionary algorithms, DRL leverages the structure of the MDP in the form of the Bellman equation to achieve better sample efficiency. The objective is to find an optimal policy $\pi_\aparam$, which maximizes the expected return or fitness $F(\pi_\aparam)$. In reinforcement learning, many approaches try to estimate the action-value function $Q^\pi(s, a) = \mathbb{E}_\pi \left[ \sum_{t=0}^{T-1} \gamma^t r_t | s, a \right]$ defined as the expected discounted return starting from state $s$, taking action $a$ and thereafter following policy $\pi$.

The Twin Delayed Deep Deterministic Policy Gradient (TD3) algorithm~\cite{fujimoto_addressing_2018} is an actor-critic, off-policy reinforcement learning method that achieves state-of-the-art results in environments with large and continuous action space. TD3 indirectly learns a policy $\pi_\aparam$ via maximization of the action-value function $Q_\cparam(s, a)$. The approach is closely connected to Q-learning~\cite{fujimoto_addressing_2018} and tries to approximate the optimal action-value function $Q^*(s, a)$ in order to find the optimal action $a^*(s) = \argmax_a Q^*(s, a)$. However, computing the maximum over action in $\max_a Q_\cparam (s, a)$ is intractable in continuous action space, so it is approximated using $\max_a Q_\cparam(s, a) = Q_\cparam(s, \pi_\aparam(s))$. In TD3, the policy $\pi_\aparam$ takes actions in the environment and the transitions are stored in a replay buffer. The collected experience is then used to train a pair of critics $Q_{{\cparam_1}}$, $Q_{{\cparam_2}}$ using temporal difference and target networks $Q_{{\cparam_1}^\prime}$, $Q_{{\cparam_2}^\prime}$ are updated to slowly track the main networks. Both critics use a single regression target $y$, calculated using whichever of the two critics gives a smaller target value and using target policy smoothing by sampling a noise $\epsilon \sim \text{clip}(\mathcal{N}(0, \sigma), -c, c)$:
\begin{equation}
\label{eq:td3-target}
y = r(s_t, a_t) + \gamma \min_{i=1,2} Q_{{\cparam_i}^\prime} \left(s_{t+1}, \pi_{\aparam^\prime}(s_{t+1}) + \epsilon\right)
\end{equation}
Both critics are learned by regression to this target and the policy is learned with a delay, only updated every $\delay$ iterations simply by maximizing $Q_{\cparam_1}$ with $\max_\aparam \mathbb{E} \left[ Q_{\cparam_1} (s, \pi_\aparam (s)) \right]$. The actor is updated using the deterministic policy gradient:
\begin{equation}
\label{eq:pg}
\nabla_\aparam J(\aparam) = \mathbb{E} \left[ \nabla_a Q_{\cparam_1}(s, a)|_{a=\pi_{\aparam}(s)} \nabla_\aparam \pi_{\aparam}(s) \right]
\end{equation}

\subsection{PGA-MAP-Elites}
\label{sec:pga-map-elites}

Policy Gradient Assisted MAP-Elites (PGA-MAP-Elites)~\cite{nilsson_policy_2021} is an extension of MAP-Elites that is designed to evolve deep neural networks by combining the directed search power and sample efficiency of DRL methods with the exploration capabilities of genetic algorithms, see Alg.~\ref{alg:pga-map-elites}. The algorithm follows the usual MAP-Elites loop of selection, variation, evaluation and addition for a budget of $I$ iterations, but uses two parallel variation operators: half of the offspring are generated using a standard Genetic Algorithm (GA) variation operator and half of the offspring are generated using a Policy Gradient (PG) variation operator. During each iteration of the loop, PGA-MAP-Elites stores the transitions from offspring evaluation in a replay buffer $\mathcal{B}$ and uses it to train a pair of critics based on the TD3 algorithm, described in Alg.~\ref{alg:train-actor-critic}. The trained critic is then used in the PG variation operator to update the selected solutions from the archive for $\npg$ gradient steps to select actions that maximize the approximated action-value function, as described in Alg.~\ref{alg:variation-pg}. At each iteration, the critics are trained for $\ncritic$ steps of gradients descents towards the target described in Eq.~\ref{eq:td3-target} averaged over $N$ transitions of experience sampled uniformly from the replay buffer $\mathcal{B}$. The actor (also named greedy actor~\cite{nilsson_policy_2021}) learns with a delay $\delay$ via maximization of the critic according to Eq.~\ref{eq:pg}.

\section{Related Work}

\subsection{Scaling QD to Neuroevolution}

The challenge of evolving diverse solutions in a high-dimensional search space has been an active research subject over the recent years.
ME-ES~\cite{colas_scaling_2020} scales MAP-Elites to high-dimensional solutions parameterized by large neural networks. This algorithm leverages Evolution Strategies to perform a directed search that is more efficient than random mutations used in Genetic Algorithms. Fitness gradients are estimated locally from many perturbed versions of the parent solution to generate a new one. The population tends towards regions of the parameter space with higher fitness but it requires to sample and evaluate a large number of solutions, making it particularly data inefficient.
In order to use the time step level information and hence improve data efficiency, methods that combine MAP-Elites with Reinforcement Learning~\cite{nilsson_policy_2021, pierrot_diversity_2022, tjanaka_approximating_2022,pierrot_evolving_2023} have emerged and proved to efficiently evolve populations of high-performing and diverse neural network for complex tasks. PGA-MAP-Elites~\cite{nilsson_policy_2021} uses policy gradients for part of its mutations, see section \ref{sec:pga-map-elites} for details. CMA-MEGA~\cite{tjanaka_approximating_2022} estimates descriptor gradients with Evolution Strategies and combines the fitness gradient and the descriptor gradients with a CMA-ES mechanism~\cite{hansen_cma_2016,fontaine_differentiable_2021}. QD-PG~\cite{pierrot_diversity_2022} introduces a diversity reward based on the novelty of the states visited and derives a policy gradient for the maximization of those diversity rewards which helps exploration in settings where the reward is sparse or deceptive. PBT-MAP-Elites~\cite{pierrot_evolving_2023} mixes MAP-Elites with a population based training process~\cite{jaderberg_population_2017} to optimize hyper-parameters of diverse RL agents.
Interestingly, recent work~\cite{tjanaka_training_2022} scales the algorithm CMA-MAE~\cite{fontaine_covariance_2023} to high-dimensional policies on robotics tasks with pure Evolution Strategies while showing comparable data efficiency to QD-RL approaches. It shows competitiveness but is still outperformed by PGA-MAP-Elites.

\subsection{Conditioning the critic}

None of the above methods takes a descriptor into account when deriving policy gradients used to mutate solutions. In other words, they do not use descriptor-conditioned policies nor descriptor-conditioned critics as our method \algonameshort{} does.
The concept of descriptor-conditioned critic is related to the concept of Universal Value Function Approximators~\cite{schaul_universal_2015} and the most related field to Quality-Diversity that uses it is Skill Discovery Reinforcement Learning~\cite{chalumeau_neuroevolution_2022}. In VIC, DIAYN, DADS, SMERL~\cite{gregor_variational_2016,eysenbach_diversity_2018,sharma_dynamics-aware_2020,kumar_one_2020}, conditioned actor-critic are used but the condition is a sampled prior and does not correspond to a real posterior like in \algonameshort{}. Furthermore, those methods use diversity at the step level and not explicitly at the trajectory level like ours. Finally, they do not use an archive to store their population, resulting in much smaller sets of final policies. Ultimately, it has been shown that behaviors evolved by QD methods are competitive with skills learned by this family of methods~\cite{chalumeau_neuroevolution_2022}, in regards to their use for adaptation and hierarchical learning.

\subsection{Archive distillation}

Distilling the knowledge of an archive into a single neural model is an alluring process that reduces the number of parameters outputted by the algorithm and enables generalization and interpolation/extrapolation. Although distillation is usually referring to policy distillation --- learning the observation/action mapping from a teacher policy --- we present archive distillation as a general term referring to any kind of knowledge transfer from an archive to another model, should it be the policies, transitions experienced in the environment, full trajectories or discovered descriptors.

To the best of our knowledge, two QD-related works use the concept of archive distillation. Go-Explore~\cite{ecoffet_go-explore_2021} stores an archive of reached states and trains a goal-conditioned policy to reproduce the trajectory of the policy that reached that state. Another interesting approach to archive distillation is to learn a generative policy network~\cite{jegorova_behavioural_2019} over the policy contained in the archive. Our approach \algonameshort{} distills the experience of the archive into a single versatile policy.

\section{\algonameshort{}}

Our new method \algonamelong{} (\algonameshort{}) overcomes limitations of PGA-MAP-Elites by leveraging a descriptor-conditioned critic to improve the PG variation operator and concurrently distills the knowledge of the archive in a single versatile policy as a by-product of the actor-critic training. The pseudocode is provided in Alg.~\ref{alg:pga-map-elites-distillation}. The algorithm follows the usual MAP-Elites loop of selection, variation, evaluation and addition for a budget of $I$ iterations. Two complementary and independent variation operators are used in parallel: 1) a standard GA operator 2) a descriptor-conditioned PG operator. At each iteration, the transitions from the evaluation step are stored in a replay buffer and used to train an actor-critic pair based on TD3.

Contrary to PGA-MAP-Elites, the actor-critic pair is descriptor-conditioned. In addition to the state $s$ and action $a$, the critic $Q_\cparam(s, a \mid d)$ also depends on the descriptor $d$ and estimates the expected discounted return starting from state $s$, taking action $a$ and thereafter following policy $\pi$ \emph{and} achieving descriptor $d$. Achieving descriptor $d$ means that the descriptor of the trajectory generated by the policy $\pi$ should have descriptor $d$. In addition to the state $s$, the actor $\pi_\aparam(s \mid d)$ also depends on the descriptor $d$ and maximizes the expected discounted return conditioned on achieving descriptor $d$. Thus, the goal of the descriptor-conditioned actor is to achieve the input descriptor $d$ \emph{while} maximizing fitness.

\begin{algorithm}
\caption{\algonameshort{}}
\label{alg:pga-map-elites-distillation}
\begin{algorithmic}
\small
\Require batch size $b$, number of GA variations $g \leq b$
\State Initialize archive $\mathcal{X}$ with $b$ random solutions and replay buffer $\mathcal{B}$
\State Initialize critic networks $Q_{\cparam_1}$, $Q_{\cparam_2}$ and actor network $\pi_\aparam$
\State $i \gets 0$
\While{$i < I$}
    \State $\textsc{train\_actor\_critic}(Q_{\cparam_1}, Q_{\cparam_2}, \pi_\aparam, \mathcal{B})$
    \State $\pi_{\sparam_1}, \dots, \pi_{\sparam_b} \gets \textsc{selection}(\mathcal{X})$
    \State $\pi_{\widehat{\sparam}_1}, \dots, \pi_{\widehat{\sparam}_g} \gets \textsc{variation\_ga}(\pi_{\sparam_1}, \dots, \pi_{\sparam_g})$
    \State $\pi_{\widehat{\sparam}_{g+1}}, \dots, \pi_{\widehat{\sparam}_b} \gets \textsc{variation\_pg}(\pi_{\sparam_{g+1}}, \dots, \pi_{\sparam_b}, Q_{\cparam_1}, \mathcal{B})$
    \State $\textsc{addition}(\pi_{\widehat{\sparam}_1}, \dots, \pi_{\widehat{\sparam}_b}, \mathcal{X}, \mathcal{B})$
    \State $i\gets i + b$  
\EndWhile

\Function{\textsc{addition}}{$\mathcal{X}, \mathcal{B}, \pi_\aparam, \pi_{\widehat{\sparam}} \dots$}
    \For{$d^\prime \in \mathcal{D}$ sampled from $b$ solutions in $\mathcal{X}$}
        \State $(f, \text{transitions}) \gets F(\pi_\aparam(\,. \mid d^\prime))$
        \State $\textsc{insert}(\mathcal{B}, \text{transitions})$
    \EndFor
    \For{$\pi_{\widehat{\sparam}} \dots$}
        \State $(f, \text{transitions}) \gets F(\pi_{\widehat{\sparam}})$, $d \gets D(\pi_{\widehat{\sparam}})$
        \State $\textsc{insert}(\mathcal{B}, \text{transitions})$
        \If{$\mathcal{X}(d) = \emptyset$ or $F(\mathcal{X}(d)) < f$}
            \State $\mathcal{X}(d) \gets \pi_{\widehat{\sparam}}$
        \EndIf
    \EndFor
\EndFunction
\end{algorithmic}
\end{algorithm}

\subsection{Descriptor-Conditioned Critic}
\label{sec:dc-critic}

Instead of estimating the action-value function with $Q_\cparam(s, a)$, we want to estimate the descriptor-conditioned action-value function with $Q_\cparam(s, a \mid d)$. When a policy $\pi$ interacts with the environment for an episode of length T, it generates a trajectory $\tau$, which is a sequence of transitions:
$$\left(s_0, a_0, r_0, s_1\right), \dots, \left(s_{T-1}, a_{T-1}, r_{T-1}, s_T\right)$$
with descriptor $D(\pi) = d$. We extend the definition of a transition $(s, a, r, s^\prime)$ to include the descriptor $d$ of the policy $(s, a, r, s^\prime, d)$. Thus, a trajectory $\tau$ with descriptor $d$ gives a sequence of transitions:
$$\left(s_0, a_0, r_0, s_1, d\right), \dots, \left(s_{T-1}, a_{T-1}, r_{T-1}, s_T, d\right)$$
However, the descriptor is only available at the end of the episode, therefore the transitions can only be augmented with the descriptor after the episode is done. In all the tasks we consider, the reward function is positive $r \colon \mathcal{S} \times \mathcal{A} \to \mathbb{R}^+$ and hence, the fitness function $F$ and action-value function are positive as well. Thus, for any sampled descriptor $d^\prime \in \mathcal{D}$, we define the descriptor-conditioned critic as equal to the normal action-value function when the policy achieves the sampled descriptor $d^\prime$ and as equal to zero when the policy does not achieve the sampled descriptor $d^\prime$. Given a transition $(s, a, r, s^\prime, d)$, and $d^\prime \in \mathcal{D}$,
\begin{equation}
\label{eq:dc-critic-1}
Q_\cparam(s, a \mid d^\prime) = 
    \begin{cases}
        Q_\cparam(s, a), & \text{if } d = d^\prime\\
        0, & \text{if } d \neq d^\prime
    \end{cases}
\end{equation}
However, with this piecewise definition, the descriptor-conditioned action-value function is not continuous and violates the universal approximation theorem continuity hypothesis~\cite{hornik_multilayer_1989}. To address this issue, we introduce a similarity function $S \colon \mathcal{D}^2 \to ]0, 1]$ defined as $S(d, d^\prime) = e^{-\frac{||d-d^\prime||_\mathcal{D}}{l}}$ to smooth the descriptor-conditioned critic and relax Eq.~\ref{eq:dc-critic-1} into:
\begin{align}
\label{eq:dc-critic-2}
Q_\cparam(s, a \mid d^\prime) =  S(d, d^\prime) \, Q_\cparam(s, a)\nonumber 
& = S(d, d^\prime) \, \mathbb{E}_\pi \left[ \sum_{t=0}^{T-1} \gamma^t r_t \Bigg\vert s, a \right]\nonumber\\
& =  \mathbb{E}_\pi \left[ \sum_{t=0}^{T-1} \gamma^t S(d, d^\prime) r_t \Bigg\vert s, a \right]
\end{align}
With Eq.~\ref{eq:dc-critic-2}, we demonstrate that learning the descriptor-conditioned critic is equivalent to scaling the reward by the similarity $S(d, d^\prime)$ between the descriptor of the trajectory $d$ and the sampled descriptor $d^\prime$. Therefore, the critic target in Eq.~\ref{eq:td3-target} is modified to include the similarity scaling and the descriptor-conditioned actor:
\begin{equation}
\label{eq:dc-td3-target}
y = S(d, d^\prime) \, r(s_t, a_t) + \gamma \min_{i=1,2} Q_{{\cparam_i}^\prime} \left(s_{t+1}, \pi_{\aparam^\prime}(s_{t+1} \mid d^\prime) + \epsilon \mid d^\prime \right)
\end{equation}
If the sampled descriptor $d^\prime$ is approximately equal to the observed descriptor $d$ of the trajectory $d \approx d^\prime$ then we have $S(d, d^\prime) \approx 1$ so the reward is unchanged. However, if the descriptor $d^\prime$ is very different from the observed descriptor $d$ then, the
reward is scaled down to $S(d, d^\prime) \, r(s_t, a_t) \approx 0$. The scaling ensures that the magnitude of the reward depends not only on the quality of the action $a$ with regards to the fitness function $F$, but also on achieving the descriptor $d^\prime$. Given one transition $(s, a, r, s^\prime, d)$, we can generate infinitely many critic updates by sampling $d^\prime \in \mathcal{D}$. This is leveraged in the new actor-critic training introduced with \algonameshort{}, which is detailed in Alg.~\ref{alg:pga-map-elites-distillation-train-actor-critic} and section \ref{sec:actor-critic-training}.

\begin{algorithm}
\caption{Descriptor-conditioned Actor-Critic Training}
\label{alg:pga-map-elites-distillation-train-actor-critic}

\begin{algorithmic}
\small
\Function{\textsc{train\_actor\_critic}}{$Q_{\cparam_1}, Q_{\cparam_2}, \pi_\aparam, \mathcal{B}$}
    \For{$t = 1  \rightarrow \ncritic$}
        \State Sample $N$ transitions $\left(s, a, r(s, a), s^\prime, d, d^\prime \right)$ from $\mathcal{B}$
        \State Sample smoothing noise $\epsilon$
        \State $y \gets S(d, d^\prime) \, r(s, a)+ \gamma \min\limits_{i=1,2}  Q_{\cparam_{i}^\prime}\left(s^\prime, \pi_{\aparam^\prime}(s^\prime \mid d^\prime) + \epsilon \mid d^\prime \right)$
        \State Update both critics by regression to $y$
        \If{$t$ mod $\delay$}
            \State Update actor using the deterministic policy gradient:
            \State 	$\frac{1}{N}\sum \nabla_a Q_{\cparam_1}(s, a \mid d^\prime)|_{a=\pi_{\aparam}(s \mid d^\prime)} \nabla_\aparam \pi_{\aparam}(s \mid d^\prime)$
            \State Soft-update target networks $Q_{\cparam{i}^\prime}$ and $\pi_{\aparam^\prime}$
        \EndIf
    \EndFor
\EndFunction
\end{algorithmic}
\end{algorithm}

\subsection{Descriptor-Conditioned Actor and Archive~Distillation}

The training of the critic requires to train an actor $\pi_\aparam$ to approximate the optimal action $a^*$ as explained in section \ref{sec:drl}. However, in this work, the action-value function estimated by the critic is conditioned on a descriptor $d$. Hence, we don't want to estimate the best action globally, but the best action given that the policy should achieve descriptor $d$ by the end of the trajectory. Therefore, the actor is extended to a descriptor-conditioned policy $\pi_\aparam(s \mid d)$, that maximizes the descriptor-conditioned critic's value with $\max_\aparam \mathbb{E} \left[ Q_\cparam (s, \pi_\aparam (s \mid d) \mid d) \right]$. The actor is updated using the deterministic policy gradient, see Alg.~\ref{alg:pga-map-elites-distillation-train-actor-critic}:
\begin{equation}
\label{eq:dc-pg-actor}
\nabla_\aparam J(\aparam) = \frac{1}{N}\sum \nabla_a Q_{\cparam_1}(s, a \mid d^\prime)|_{a=\pi_{\aparam}(s \mid d^\prime)} \nabla_\aparam \pi_{\aparam}(s \mid d^\prime)
\end{equation}
The policy $\pi_\aparam(s \mid d)$ learns to suggest actions $a$ that optimize reward $r(s, a)$ \emph{while} generating a trajectory achieving descriptor $d$. Consequently, the descriptor-conditioned actor can exhibit a wide range of descriptors, effectively distilling some of the capabilities of the archive into a single versatile policy.

\subsection{Actor-Critic Training}
\label{sec:actor-critic-training}

In section \ref{sec:dc-critic}, we show that the new descriptor-conditioned critic target in Eq.~\ref{eq:dc-td3-target} requires a sampled descriptor $d^\prime$. At each iteration, the transitions $(s, a , r(s, a), s^\prime, d)$ generated from the evaluation of offspring are stored in a replay buffer $\mathcal{B}$. To learn the relation between being in state $s$, taking action $a$ and achieving descriptor $d^\prime$, the critic needs to be trained on ``positive'' samples i.e. samples with $d^\prime = d$ and ``negative'' samples i.e. samples with $d^\prime \neq d$. Positive samples are easy to generate because we can sample $d^\prime$ as equal to the observed descriptor $d$ of the trajectory. However, negative samples require to define an arbitrary sampling strategy that can impact the performance of the algorithm.

Moreover, learning to act from data without active interactions in the environment is difficult~\cite{ostrovski_difficulty_2021}. If the actor only learns passively on rollouts performed by offspring from policies stored in the archive, strong deviation from the state-action distribution can happen, preventing convergence. To solve this passive learning problem and generate negative samples at the same time, we evaluate the actor $\pi_\aparam$ in the environment. At each iteration, we sample a batch of $b$ descriptors $d^\prime$ in $\mathcal{D}$ and evaluate $\pi_\aparam(\,. \mid d^\prime)$. The actor evaluation step gives $b$ trajectories with transitions $(s_t, a_t, r_t, s_{t+1}, d, d^\prime)$ with $d$ the observed descriptor achieved by the trajectory and $d^\prime$ the input descriptor that is sampled for evaluation. In general, the observed descriptor and the sampled descriptor are different, $d^\prime \neq d$, providing negative samples to the critic training. To get the descriptors $d^\prime$, we uniformly sample with replacement $b$ solutions $\pi_\sparam$ from the archive and take their descriptors with an additional Gaussian noise. The solutions have descriptors $D(\pi_\sparam) = d_\sparam$ and we compute a batch of descriptors $d^\prime = d_\sparam + \mathcal{N}(0_\mathcal{D}, \sigma_d I)$. The negative samples from the actor evaluation are stored in the replay buffer alongside the positive samples from offspring evaluation for which we take $d^\prime = d$, as detailed in the \textsc{addition} function of Alg.~\ref{alg:pga-map-elites-distillation}.

\subsection{Descriptor-Conditioned PG variation}

Once the critic $Q_\cparam(s, a \mid d)$ is learned, it can be used to improve the fitness of solutions in the archive, as described in Alg.~\ref{alg:variation-pg-distillation}. A parent solution $\pi_\sparam$ with descriptor $D(\pi_\sparam) = d_\sparam$ is selected from the archive. Then, we apply the PG variation operator, using the descriptor $d_\sparam$ to condition the critic, and thus, to apply $\npg$ descriptor-conditioned gradient steps using the deterministic policy gradient:
\begin{equation}
\label{eq:dc-pg}
\nabla_\sparam J(\sparam) = \frac{1}{N}\sum \nabla_a Q_{\cparam_1}(s, a \mid d_\sparam)|_{a=\pi_{\sparam}(s)} \nabla_\sparam \pi_{\sparam}(s)
\end{equation}
The goal is to improve the fitness of the solution while remaining in the same cell as the parent.

\begin{algorithm}
\caption{Descriptor-conditioned PG Variation}
\label{alg:variation-pg-distillation}
\begin{algorithmic}
\small
\Function{\textsc{variation\_pg}}{$\pi_{\sparam} \dots,  Q_{\cparam_1}, \mathcal{B}$}
    \For{$\pi_{\sparam} \dots$}
        \State $d_\sparam \gets D(\pi_{\sparam})$
        \For{$i = 1  \rightarrow \npg$}
            \State Sample $N$ transitions $\left(s, a, r(s, a), s^\prime, d, d^\prime \right)$ from $\mathcal{B}$
            \State Update actor using the deterministic policy gradient:
            \State 	$\frac{1}{N}\sum \nabla_a Q_{\cparam_1}(s, a \mid d_\sparam)|_{a=\pi_{\sparam}(s)} \nabla_\sparam \pi_{\sparam}(s)$
        \EndFor
    \EndFor
    \State \Return $\pi_{\widehat{\aparam}} \dots$
\EndFunction
\end{algorithmic}
\end{algorithm}

\section{Experiments}

\subsection{Evaluation Tasks}

We evaluate \algonameshort{} on four QDGym tasks~\cite{nilsson_policy_2021} implemented in Brax~\cite{freeman_brax_2021} and derived from standard DRL benchmarks for robot locomotion. We call these tasks ``Ant-Omni'', ``AntTrap-Omni'', ``Hexapod-Omni'', and ``Walker-Uni''. Three omnidirectional tasks and one unidirectional task are considered for evaluation. Ant-Omni, AntTrap-Omni and Hexapod-Omni are omnidirectional tasks in which the objective is for the robot to reach all possible points in the plane while minimizing energy consumption. Walker-Uni is a unidirectional task in which the objective is for the robot to discover all possible ways to walk forward while maximizing a trade-off between speed and energy consumption. Walker-Uni is exactly the same tasks used in PGA-MAP-Elites paper~\cite{nilsson_policy_2021}, Ant-Omni and Hexapod-Omni have been used by Flageat and Lim et al.~\cite{flageat_benchmarking_2022} and AntTrap-Omni is adapted from QD-PG paper~\cite{pierrot_diversity_2022}. The difference is the elimination of the forward reward term in the reward function. The goal of AntTrap-Omni is to reach every point in the plane while minimizing energy consumption in a deceptive environment. This new task is designed to show that our algorithm performs well in environment where there is a discontinuity on the fitness landscape in descriptor space caused by the trap. The trap is causing a discontinuity that the descriptor-conditioned critic has to learn. Indeed, points on both sides of the trap will be close in the descriptor space, but distant in terms of trajectory to achieve these descriptors. We detail these tasks in Tab.~\ref{tab:tasks}.

PGA-MAP-Elites has previously shown state-of-the-art results on some of these tasks, in particular Walker-Uni but tends to struggle in omnidirectional tasks. In those tasks, the global optimal of the fitness function is a solution that prevents the robot to move, which is directly opposed to the objective of discovering how to reach different locations. Hence, most of the offspring generated by the PG variation will tend to move less and travel a shorter distance. Instead, \algonameshort{} aims to improve the energy consumption while maintaining the ability to reach distant locations.

\setlength{\tabcolsep}{3pt}
\begin{table}[h]
\small
\caption{Evaluation Tasks}
\label{tab:tasks}
\centering
\newdimen\length
\length=1.4cm
\begin{tabular}{l | c c c c}
    \toprule
    & \textsc{Ant} & \textsc{AntTrap} & \textsc{Hexapod} & \textsc{Walker}\\
    & \includegraphics[height=0.8\length, width=\length]{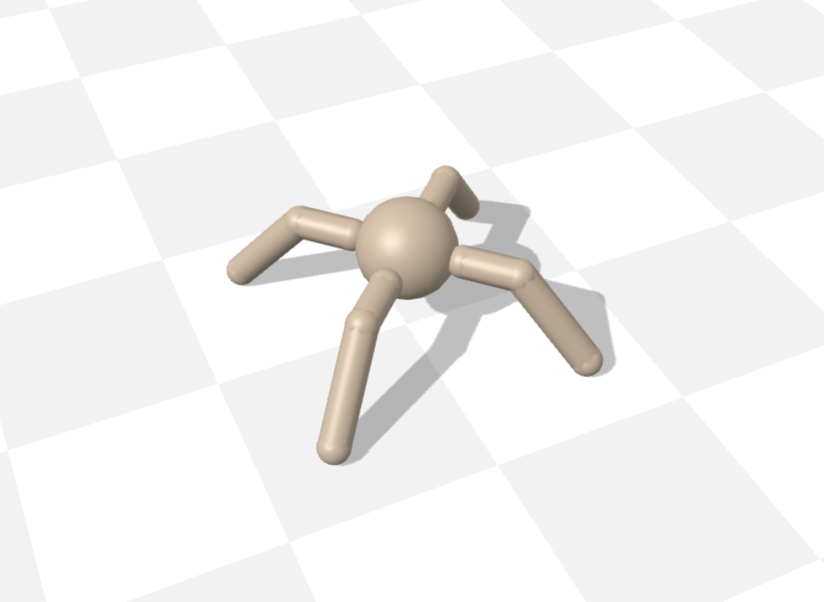} & \includegraphics[height=0.8\length, width=\length]{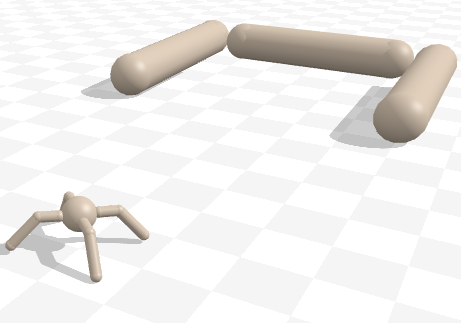} & \includegraphics[height=0.8\length, width=\length]{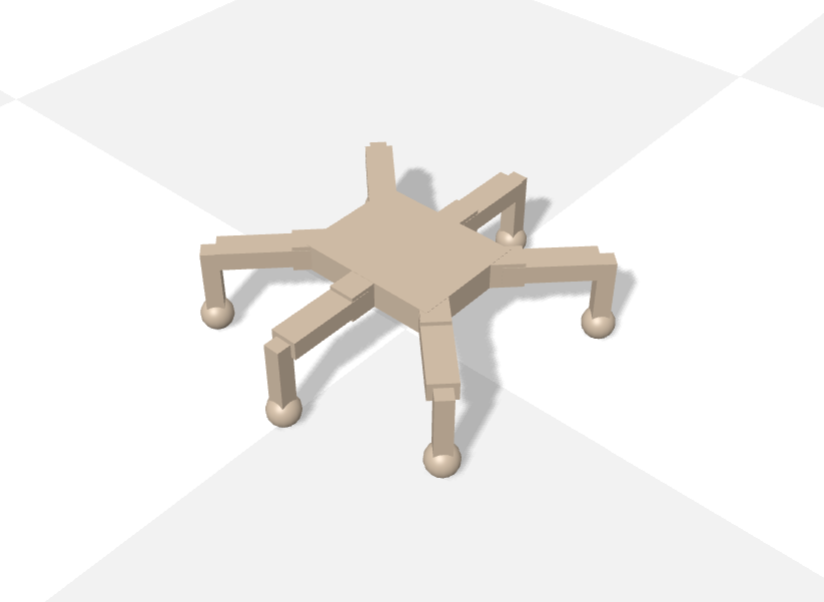} & \includegraphics[height=0.8\length, width=\length]{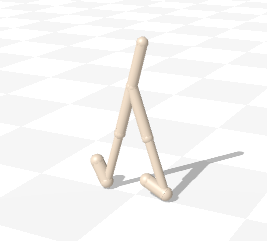}\\
    \midrule
    \textsc{State} & \multicolumn{4}{c}{Position and velocity of body and joints}\\
    \textsc{Action} & \multicolumn{4}{c}{Torques for each joint}\\
    \textsc{Type} & Omni & Omni & Omni & Uni\\
    \textsc{Descriptor} & \multicolumn{3}{c}{Final $(x, y)$ position} & Feet contact\\
    \textsc{Fitness} & \multicolumn{4}{c}{Surviving reward + Energy consumption penalty}\\
    \textsc{State dim} & 29 & 29 & 48 & 19\\
    \textsc{Action dim} & 8 & 8 & 18 & 6\\
    \textsc{Episode len} & 250 & 250 & 250 & 1000\\
    \textsc{Parameters} & 21,384 & 21,384 & 25,106 & 19,846\\
    \bottomrule
\end{tabular}
\end{table}

\subsection{Baselines}

We compare \algonameshort{} with four algorithms: CVT-MAP-Elites~\cite{vassiliades_using_2017}, MAP-Elites-ES~\cite{colas_scaling_2020}, PGA-MAP-Elites~\cite{nilsson_policy_2021} and QD-PG~\cite{pierrot_diversity_2022}. We could not compare to PBT-MAP-Elites~\cite{pierrot_evolving_2023} as its implementation is not yet available. We use QDax implementations and each experiment is replicated 20 times with random seeds, over one million evaluations. The source code of \algonameshort{} is available at \href{https://github.com/adaptive-intelligent-robotics/DCG-MAP-Elites}{github.com/adaptive-intelligent-robotics/DCG-MAP-Elites} including a container in which all the experiments can be replicated.

\subsection{Evaluation Metrics}
\label{sec:evaluation-metrics}

We consider three main metrics to evaluate the archives and two additional metrics to evaluate the descriptor-conditioned actor. We compute p-values based on the Wilcoxon rank-sum test with a Bonferroni correction for all metrics.
\begin{itemize}[leftmargin=*]
    \item \textbf{QD-score:} The sum of fitness of all solutions stored in the archive. In the task considered, fitnesses are always positive, which avoid penalizing algorithms for finding additional solutions. This score encompasses both the quality and diversity of the population.
    \item \textbf{Coverage:} The proportion of filled cells in the archive, indicating the success of descriptor space illumination.
    \item \textbf{Maximum fitness:} The fitness of the best solution in the archive.
\end{itemize}

We also consider two additional metrics to evaluate the descriptor-conditioned policy. The purpose of these metrics is to quantify how similar the archive and the descriptor-conditioned policy are in terms of QD-score and coverage of the descriptor space.
\begin{itemize}[leftmargin=*]
    \item \textbf{Descriptor-conditioned QD-score:} The sum of fitness achieved by the descriptor-conditioned policy evaluated on each descriptor of the solutions in the archive.
    $$\sum_{\pi_\sparam \in \mathcal{X}} F(\pi_\aparam(\, . \mid D(\pi_\sparam)))$$
    \item \textbf{Descriptor error mean:} (i) for the archive, the average difference over all solutions, between the stored descriptor and the reevaluated descriptor, (ii) for the descriptor-conditioned policy, the average difference over all solutions, between the stored descriptor and the reevaluated descriptor achieved by the policy conditioned on the stored descriptor. ($|\mathcal{X}|$ number of solutions in the archive).
    $$\frac{1}{|\mathcal{X}|} \sum_{\pi_\sparam \in \mathcal{X}} || D(\pi_\aparam(\, . \mid D(\pi_\sparam))) - D(\pi_\sparam)||_\mathcal{D}$$
\end{itemize}

\subsection{Results}

\subsubsection{Archive}

\begin{figure*}[ht]
\includegraphics[width=0.9\textwidth]{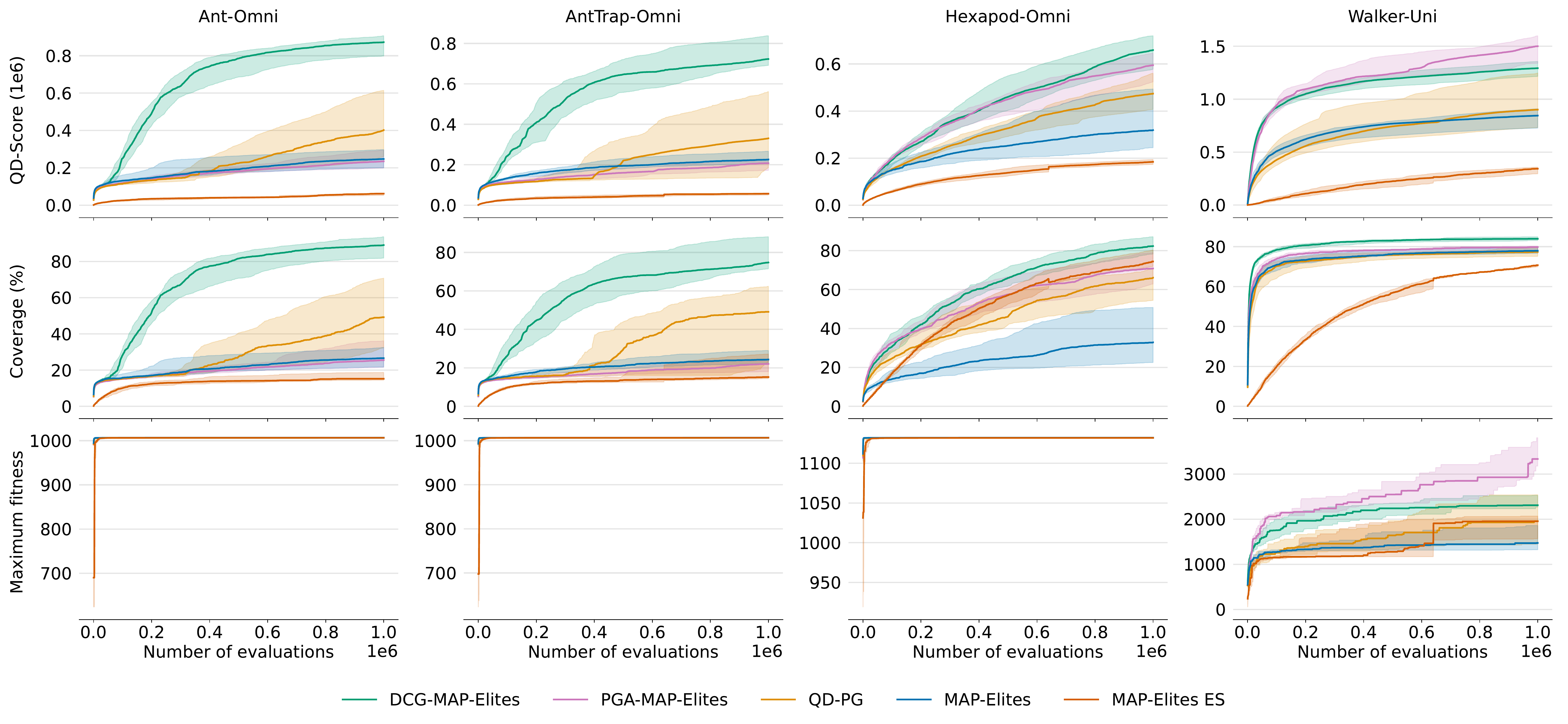}
\caption{QD-score, coverage and maximum fitness for \algonameshort{} and the baselines on all tasks. Each experiment is replicated 20 times with different random seeds. The solid line is the median and the shaded area is the first and third quartiles.}
\label{fig:plot_baseline}
\end{figure*}

The experimental results on Fig.~\ref{fig:plot_baseline} demonstrate that \algonameshort{} achieves the best archive in the two Ant tasks in terms of QD-score ($p < 10^{ -6}$) and coverage ($p < 10^{ -4}$) compared to all other baselines. On the two other tasks, the QD-score of our algorithm is similar to PGA-MAP-Elites, the previous state-of-the-art, while achieving a significantly better coverage than all baselines ($p < 10^{ -3}$). The coverage metric shows that \algonameshort{} surpasses the exploration capabilities of QD-PG on all tasks ($p < 10^{-4}$). We also show that our method still benefits from the exploration power of the GA operator even in deceptive environment like AntTrap-Omni.

\begin{figure}[H]
    \centering
    \includegraphics[width=\hsize]{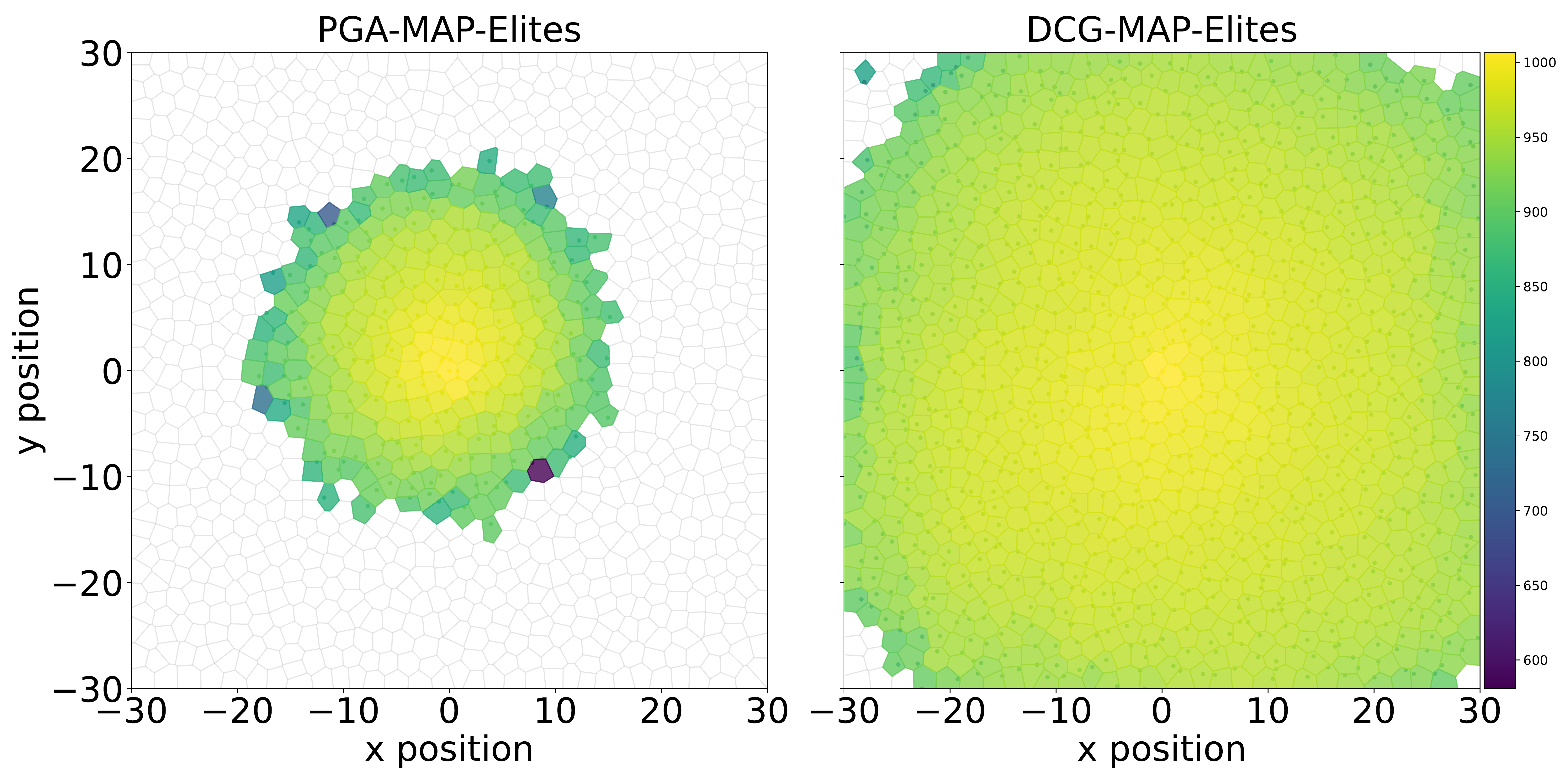}
    \caption{Final archives for PGA-MAP-Elites and \algonameshort{} on Ant-Omni after one million evaluations.}
    \label{fig:plot_archive}
\end{figure}

The experimental results confirm that \algonameshort{} is able to overcome the limits of PGA-MAP-Elites on omnidirectional tasks while still performing on par when evaluated on the unidirectional task (Walker-Uni) where no significant improvement was expected. Thus confirming the interest of having a descriptor-conditioned gradient to make the PG variation operator fruitful in a wider range of tasks. Overall, \algonameshort{} shows best (or competitive) performance on all metrics and tasks, hence proving to be the first successful effort in the QD-RL literature to achieve well on both the unidirectional and omnidirectional tasks. Previous efforts were usually adapted to either one or the other~\cite{nilsson_policy_2021,pierrot_diversity_2022,tjanaka_approximating_2022}. Nevertheless, our results show that there is still room for improvement. In particular, we can see that \algonameshort{} was not able to outperform the baselines with the same margin on Hexapod-Omni that it has been able to do on the other omnidirectional tasks. We hypothesize that the environment dynamics and the higher-dimensional search space of Hexapod-Omni makes it much more challenging.

\subsubsection{Descriptor-Conditioned Policy}


\vspace{-10pt}\begin{table}[H]
\small
\caption{Median of the QD-score and Descriptor Error Mean (DEM) after one million evaluations for the archive and for the descriptor-conditioned policy over 20 replications.}
\label{tab:descriptor-conditioned-policy}
\vspace{-10pt}\begin{tabular}{l l | c c c c}
\toprule
\textsc{Metric} & \textsc{Type} & \textsc{Ant} & \textsc{AntTrap} & \textsc{Hexapod} & \textsc{Walker}\\
\midrule
\multirow{2}{*}{\makecell{\textsc{QD-score} \\ ($10^5$)}} & Archive & $8.7$ & $7.2$ & $6.5$ & $9.8$\\
& Policy & $8.3$ & $7.0$ & $0.96$ & $5.2$\\
\hline
\multirow{2}{*}{\textsc{DEM}} & Archive & $4.04$ & $3.44$ & $0.078$ & $0.12$\\
& Policy & $6.25$ & $5.8$ & $1.39$ & $0.27$\\
\bottomrule
\end{tabular}
\end{table}

\begin{figure*}[ht]
\includegraphics[width=0.9\textwidth]{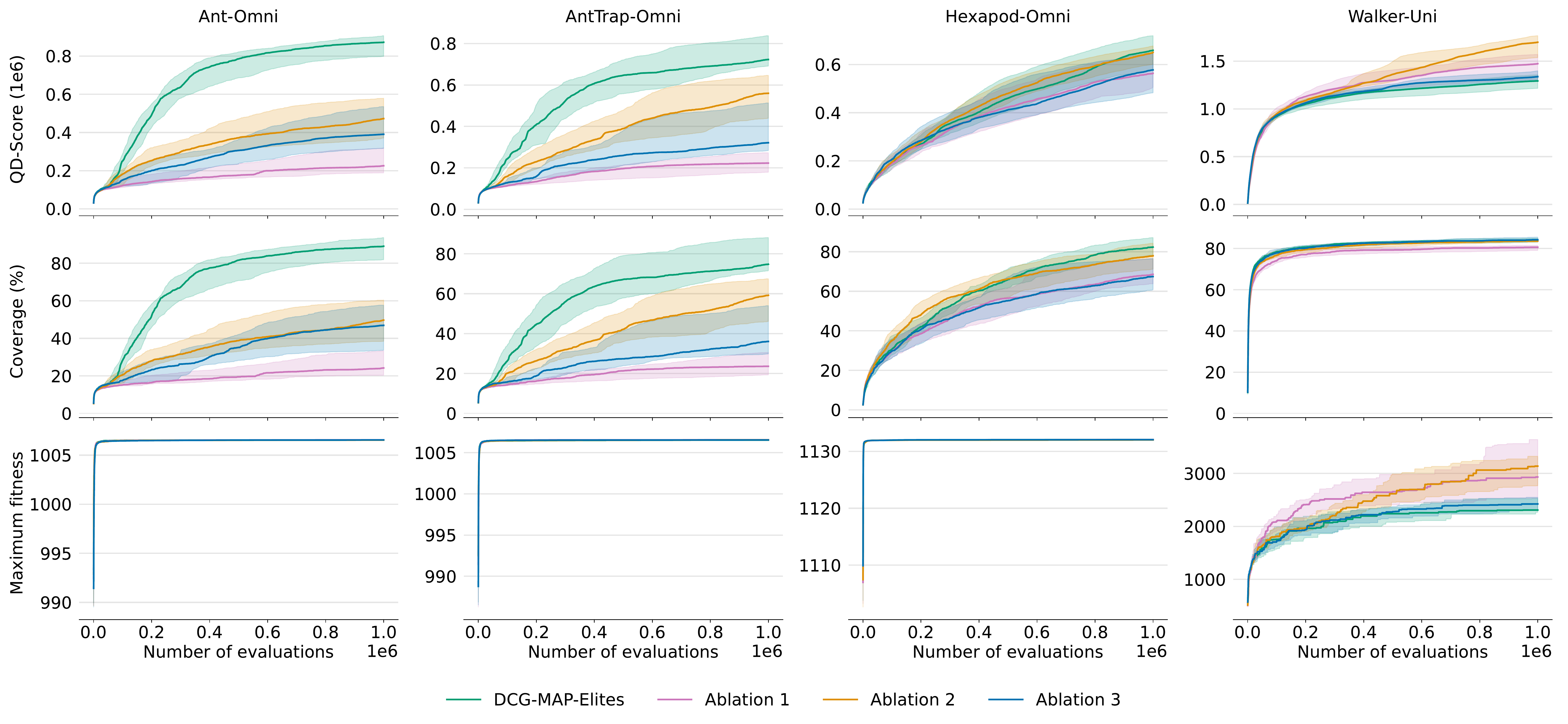}
\caption{QD-score, coverage and maximum fitness for \algonameshort{} and the ablations on all tasks. Each experiment is replicated 20 times with different random seeds. The solid line is the median and the shaded area is the first and third quartiles. Ablation (1) is \algonameshort{} without actor evaluation and without negative samples, Ablation (2) is \algonameshort{} without actor evaluation but with negative samples, Ablation (3) is \algonameshort{} without a descriptor-conditioned actor.}
\label{fig:archive-performance}
\end{figure*}

In Tab.~\ref{tab:descriptor-conditioned-policy} we provide the final QD-score and Descriptor Error Mean (DEM) of the archive and the descriptor-conditioned policy, see section \ref{sec:evaluation-metrics} for definitions. First, on Ant-Omni and AntTrap-Omni, the descriptor-conditioned policy achieves the same QD-score as the archive, measured by the descriptor-conditioned QD-score (decreases by less than 3\%). This shows that the single-conditioned policy is able to restore the quality of the archive although having compressed the information in a single network.
Second, it is interesting to cross this observation with the fact that, at the same time, the Descriptor Error Mean obtained by the conditioned policy is reasonable. We observe a DEM of 6.25, which is less than 8\% of the typical size of the environment (i.e. the diagonal of the square considered). The reader can refer to Fig.~\ref{fig:plot_archive} to get an idea of what that represents. In addition, the DEM of the archive is of 4.04 (4.7\% of size of the environment), showing an inevitable stochasticity in the evaluation that is related to the simulator used~\cite{freeman_brax_2021}. Those two combined observations (QD-score and DEM) tend to show that our archive and our descriptor-conditioned policy have similar properties on Ant-Omni and AntTrap-Omni and hence that the versatile policy could be a practical summary of our archive.
Third, we can observe that on the Hexapod-Omni and Walker-Uni tasks, the descriptor-conditioned policy struggles to recover the QD-score of the archive (reduced by more than 40\% in both cases). On the Hexapod-Omni, the DEM of the descriptor-conditioned policy is significantly larger than the typical error obtained by the archive (1.39 vs 0.078). This task is more challenging than Ant-Omni because of a higher number of dimensions (see Tab.~\ref{tab:tasks}) and we hypothesize that the optimization algorithm used (TD3) could be the limitation. For Walker-Uni, the DEM of the archive (0.12) shows that there is also an inherent variability of the descriptor, we are hence less surprised by the DEM obtained by the single-conditioned policy (0.27) although it could make it less usable in practice.

Overall, those results shows that our single descriptor-conditioned policy can already be seen as a promising summary of our archive, showing very similar properties on half our tasks. The other tasks prove that there is still room for improvement of our method.

\subsubsection{Ablations}

We perform additional ablation experiments to show the importance of three components of \algonameshort{}, namely: using negative samples for the critic's training, actively collecting experiences with the actor, and conditioning the actor on the descriptor.
In \textbf{Ablation (1)}, we remove the actor evaluation in the environment. This allows to study the impact of both passive learning and of lack of negative samples, see section \ref{sec:actor-critic-training} for a detailed explanation. The lack of active learning deprives the learning of negative samples that are necessary to learn the descriptor-conditioned action-value function. We observe that ablation (1) is performing worse than \algonameshort{} on all omnidirectional tasks ($p < 0.01$).
However it performs similarly on Walker-Uni, which can be explained by the fact that non-conditioned PG emitter already prove efficient in solving this task.
This first ablation result highlight the importance of active learning and negative examples for the critic's training.
To complement this result, in \textbf{Ablation (2)}, we remove the actor evaluation but we generate negative samples artificially by sampling in the descriptor space. We show that ablation (2) is performing significantly worse than \algonameshort{} on Ant tasks ($p < 10^{-3}$) but performs similarly on Hexapod-Omni and Walker-Uni. However, ablation (2) is performing significantly better than ablation (1) on Ant tasks ($p < 10^{-5}$), demonstrating the importance of both active learning \emph{and} negative samples.
Finally, in \textbf{Ablation (3)} we replace the descriptor-conditioned actor $\pi_\aparam(\,. \mid d)$ with a normal actor $\pi_\aparam(\,. \,)$. Ablation (3) is performing significantly worse on all omnidirectional tasks ($p < 0.05$), demonstrating the importance of the descriptor-conditioned actor for the critic's training. Ablation (3) performs equivalently to \algonameshort{} on Walker-Uni.

\section{Conclusion}

In this work, we introduce \algonameshort{} and demonstrate the benefits of having descriptor-conditioned gradients to evolve populations of large neural networks. We concurrently train a descriptor-conditioned actor, as a by-product of the critic's training, that can achieve a wide range of descriptors. Our method performs better than PGA-MAP-Elites by a significant margin in omnidirectional tasks while maintaining similar performance to PGA-MAP-Elites in unidirectional tasks. We also show that the synergy between the fitness improvement capabilities of the PG variations and the exploration capabilities of the GA variations is preserved, even in deceptive environments. On some tasks, the descriptor-conditioned policy demonstrates properties that are similar to the discrete archive, summarizing its capabilities into one single neural network and acting as a continuous archive. We think that distilling the archive into a single policy is a promising method as it enables to have less redundancy compared to a discrete archive in which most of the solutions can be similar, especially between close cells. The descriptor-conditioned policy could also negate the burden of dealing with archive of thousands of solutions in practical applications.

The benefits of combining DRL methods with MAP-Elites come with the limitations of MDP settings. Specifically, we are limited to evolving differentiable solutions and the foundations of DRL algorithms rely on the Markov property and full observability. In this work in particular, we face challenges with the Markov property because the descriptors depend on full trajectories. Thus, the scaled reward introduced in our method depends on the full trajectory and not only on the current state and action. The performance of the descriptor-conditioned policy also shows that there is room for improvement to better distill the knowledge of the archive.

For future work, we would like to investigate the generalization capabilities of the descriptor-conditioned policy trained with \algonameshort{} and try to produce solutions with descriptors that are not in the archive, effectively performing descriptor space interpolation. In our method, the critic attempts to mutate solutions to produce offspring with higher fitness while keeping their descriptors constant. We think that we could use the descriptor-conditioned critic to mutate solutions to produce offspring towards different descriptors, thereby explicitly promoting diversity.

\bibliographystyle{ACM-Reference-Format}
\bibliography{bibliography}

\clearpage
\appendix

\section*{Supplementary material --- MAP-Elites with Descriptor-Conditioned Gradients and Archive~Distillation into a Single Policy}

\section{Algorithms}

\subsection{MAP-Elites}

\begin{algorithm}[H]
\caption{MAP-Elites}
\label{alg:map-elites}

\begin{algorithmic}
\small
\Require batch size $b$
\State Initialize archive $\mathcal{X}$ with $b$ random solutions
\State $i \gets 0$
\While{$i < I$}
    \State $x_1, \dots, x_b \gets \textsc{selection}(\mathcal{X})$
    \State $\widehat{x}_1, \dots, \widehat{x}_b \gets \textsc{variation}(x_1,\dots, x_b)$
    \State $\textsc{addition}(\mathcal{X}, \widehat{x}_1, \dots, \widehat{x}_b)$
    \State $i \gets i + b$
\EndWhile

\Function{addition}{$\mathcal{X}, \widehat{x}\dots$} :
    \For{$\widehat{x} \dots$}
        \State $f \gets F(\widehat{x})$, $d \gets D(\widehat{x})$
        \If{$\mathcal{X}(d) = \emptyset$ or $F(\mathcal{X}(d)) < f$}
            \State $\mathcal{X}(d) \gets \widehat{x}$
        \EndIf
    \EndFor
\EndFunction
\end{algorithmic}
\end{algorithm}

\subsection{PGA-MAP-Elites}

\begin{algorithm}[H]
\caption{PGA-MAP-Elites}
\label{alg:pga-map-elites}
\begin{algorithmic}
\small
\Require batch size $b$, number of GA variations $g \leq b$
\State Initialize archive $\mathcal{X}$ with $b$ random solutions and replay buffer $\mathcal{B}$
\State Initialize critic networks $Q_{\cparam_1}$, $Q_{\cparam_2}$ and actor network $\pi_\aparam$
\State $i \gets 0$
\While{$i < I$}
    \State $\textsc{train\_actor\_critic}(Q_{\cparam_1}, Q_{\cparam_2}, \pi_\aparam, \mathcal{B})$
    \State $\pi_{\sparam_1}, \dots, \pi_{\sparam_{b-1}} \gets \textsc{selection}(\mathcal{X})$
    \State $\pi_{\widehat{\sparam}_1}, \dots, \pi_{\widehat{\sparam}_g} \gets \textsc{variation\_ga}(\pi_{\sparam_1}, \dots, \pi_{\sparam_g})$
    \State $\pi_{\widehat{\sparam}_{g+1}}, \dots, \pi_{\widehat{\sparam}_{b-1}} \gets \textsc{variation\_pg}(\pi_{\sparam_{g+1}}, \dots, \pi_{\sparam_{b-1}}, Q_{\cparam_1}, \mathcal{B})$
    \State $\textsc{addition}(\pi_{\widehat{\sparam}_1}, \dots, \pi_{\widehat{\sparam}_{b-1}}, \pi_\aparam, \mathcal{X}, \mathcal{B})$
    \State $i\gets i + b$  
\EndWhile

\Function{\textsc{addition}}{$\mathcal{X}, \mathcal{B}, \pi_{\widehat{\sparam}} \dots$}
    \For{$\pi_{\widehat{\sparam}} \dots$}
        \State $(f, \text{transitions}) \gets F(\pi_{\widehat{\sparam}})$, $d \gets D(\pi_{\widehat{\sparam}})$
        \State $\textsc{insert}(\mathcal{B}, \text{transitions})$
        \If{$\mathcal{X}(d) = \emptyset$ or $F(\mathcal{X}(d)) < f$}
            \State $\mathcal{X}(d) \gets \pi_{\widehat{\sparam}}$
        \EndIf
    \EndFor
\EndFunction
\end{algorithmic}
\end{algorithm}

\begin{algorithm}[H]
\caption{Actor-Critic Training}
\label{alg:train-actor-critic}
\begin{algorithmic}
\small
\Function{\textsc{train\_actor\_critic}}{$Q_{\cparam_1}, Q_{\cparam_2}, \pi_\aparam, \mathcal{B}$}
    \For{$t = 1  \rightarrow \ncritic{}$}
        \State Sample $N$ transitions $\left(s, a, r(s, a), s^\prime \right)$ from $\mathcal{B}$
        \State Sample smoothing noise $\epsilon$
        \State $y \gets r(s, a)+ \gamma \min\limits_{i=1,2}  Q_{\cparam_{i}^\prime}\left(s^\prime, \pi_{\aparam^\prime}(s^\prime) + \epsilon\right)$
        \State Update both critics by regression to $y$
        \If{$t$ mod $\delay$}
            \State Update actor using the deterministic policy gradient:
            \State 	$\frac{1}{N}\sum \nabla_a Q_{\cparam_1}(s, a)|_{a=\pi_{\aparam}(s)} \nabla_\aparam \pi_{\aparam}(s)$
            \State Soft-update target networks $Q_{\cparam{i}^\prime}$ and $\pi_{\aparam^\prime}$
        \EndIf
    \EndFor
\EndFunction
\end{algorithmic}
\end{algorithm}

\begin{algorithm}[H]
\caption{PG Variation}
\label{alg:variation-pg}
\begin{algorithmic}
\small
\Function{\textsc{variation\_pg}}{$\pi_{\sparam} \dots,  Q_{\cparam_1}, \mathcal{B}$}
    \For{$\pi_{\sparam} \dots$}
        \For{$i = 1  \rightarrow \npg$}
            \State Sample $N$ transitions $\left(s, a, r(s, a), s^\prime \right)$ from $\mathcal{B}$
            \State Update actor using the deterministic policy gradient:
            \State 	$\frac{1}{N}\sum \nabla_a Q_{\cparam_1}(s, a)|_{a=\pi_{\sparam}(s)} \nabla_\sparam \pi_{\sparam}(s)$
        \EndFor
    \EndFor
    \State \Return $\pi_{\widehat{\sparam}} \dots$
\EndFunction
\end{algorithmic}
\end{algorithm}

\section{Hyperparameters}

In Tab.~\ref{tab:hyperparameters} we provide the hyperparameters used for \algonameshort{} and the baselines on all tasks.

\textsc{\begin{table*}
  \caption{Hyperparameters for \algonameshort{} and all baselines}
  \label{tab:hyperparameters}
  \begin{tabular}{l | c c c c c}
    \toprule
    Parameter & MAP-Elites & MAP-Elites ES & PGA-MAP-Elites & QD-PG & DCG-MAP-Elites\\
    \midrule
    Number of centroids & $1024$ & $1024$ & $1024$ & $1024$ & $1024$\\
    Evaluation batch size $b$ & $256$ & $1050$ & $256$ & $256$ & $256$\\
    Policy networks & [128, 128, $|\mathcal{A}|$] & [128, 128, $|\mathcal{A}|$] & [128, 128, $|\mathcal{A}|$] & [128, 128, $|\mathcal{A}|$] & [128, 128, $|\mathcal{A}|$]\\
    Number of GA variations $g$ & 256 & 0 & 128 & 86 & 128\\
    \hline
    GA variation param. 1 $\sigma_1$ & $0.005$ &  & $0.005$ & $0.005$ & $0.005$\\
    GA variation param. 2 $\sigma_2$ & $0.05$ &  & $0.05$ & $0.05$ & $0.05$\\
    \hline
    Actor network &  &  & [256, 256, $|\mathcal{A}|$] & [256, 256, $|\mathcal{A}|$] & [256, 256, $|\mathcal{A}|$]\\
    Critic network &  &  & [256, 256, 1] & [256, 256, 1] & [256, 256, 1]\\
    TD3 batch size $N$ &  &  & $100$ & $100$ & $100$\\
    Quality critic training steps $\ncritic$ &  &  & $3000$ & $3000$ & $3000$\\
    Diversity critic training steps $\ncritic$ &  &  &  & $300$ & \\
    PG training steps $\npg$ &  &  & $150$ & $150$ & $150$\\
    Policy learning rate &  &  & $5 \times 10^{-3}$ & $5 \times 10^{-3}$ & $5 \times 10^{-3}$\\
    Actor learning rate &  &  & $3 \times 10^{-4}$ & $3 \times 10^{-4}$ & $3 \times 10^{-4}$\\
    Critic learning rate &  &  & $3 \times 10^{-4}$ & $3 \times 10^{-4}$ & $3 \times 10^{-4}$\\
    Replay buffer size &  &  & $10^6$ & $10^6$ & $10^6$\\
    Discount factor $\gamma$ &  &  & $0.99$ & $0.99$ & $0.99$\\
    Actor delay $\delay$ &  &  & $2$ & $2$ & $2$\\
    Target update rate &  &  & $0.005$ & $0.005$ & $0.005$\\
    Smoothing noise var. $\sigma$ &  &  & $0.2$ & $0.2$ & $0.2$\\
    Smoothing noise clip &  &  & $0.5$ & $0.5$ & $0.5$\\
    \hline
    Sample number &  & 1000 &  &  &\\
    Sample sigma &  & 0.02 &  &  &\\
    \hline
    lengthscale $l$ &  &  &  &  & 0.008\\
    Descriptor sigma $\sigma_d$ &  &  &  &  & 0.0004\\
  \bottomrule
\end{tabular}
\end{table*}}

\end{document}